\documentclass[10pt,twocolumn,letterpaper]{article}

\usepackage[pagenumbers]{cvpr} %

\DeclareMathOperator*{\argmax}{argmax}
\DeclareMathOperator*{\argmin}{argmin}

\DeclareMathOperator{\depth}{Depth}
\DeclareMathOperator{\leaves}{Leaves}

\usepackage{mathtools}

\usepackage[linesnumbered, ruled, vlined]{algorithm2e}
\SetKwInput{KwData}{In}
\SetKwInput{KwResult}{Out}
\SetArgSty{textnormal}

\usepackage{pgfplots}
\pgfplotsset{compat=1.18}
\usepgfplotslibrary{groupplots}
\pgfplotsset{
nodes near coords black white/.style={
    small value/.style={
        font=\scriptsize,
        anchor=center,
        text=white,
        xshift=-0.09cm,        
    },
    large value/.style={
        font=\scriptsize,
        anchor=center,
        text=black,
        xshift=-0.09cm,
    },
    every node near coord/.style={
        check for zero/.code={
            \pgfmathfloatifflags{\pgfplotspointmeta}{0}{
            \pgfkeys{/tikz/coordinate}
            }{
            \begingroup
            \pgfkeys{/pgf/fpu}
            \pgfmathparse{\pgfplotspointmetatransformed<700}
            \global\let\result=\pgfmathresult
            \endgroup
            \pgfmathfloatcreate{1}{1.0}{0}
            \let\ONE=\pgfmathresult
            \ifx\result\ONE
                \pgfkeysalso{/pgfplots/small value}
            \else
                \pgfkeysalso{/pgfplots/large value}
            \fi
            }
        },
        check for zero,
    },
},
hdist style/.style={  
    every tick label/.append style={font=\footnotesize},
    nodes near coords={
        \begingroup
        \pgfkeys{/pgf/fpu}
        \pgfmathparse{
            \myvalue > 10000 ? round(\myvalue / 1000) : (\myvalue > 1000 ? \myvalue / 1000 : \myvalue)
        }
        \global\let\result=\pgfmathresult
        \pgfmathparse{\myvalue > 10000}
        \global\let\nodegttenk=\pgfmathresult
        \pgfmathparse{\myvalue > 1000}
        \global\let\nodegtk=\pgfmathresult        
        \endgroup

        \pgfmathfloatifflags{\nodegttenk}{1}{
            \pgfmathprintnumber{\result}k%
        }{
            \pgfmathfloatifflags{\nodegtk}{1}{
                \pgfmathprintnumber[fixed, precision=1]{\result}k%
            }{
                \pgfmathprintnumber{\result}%
            }
        }
    },    
    visualization depends on={\thisrow{value} \as \myvalue},
    nodes near coords black white,
},
hdist base/.style={
    colormap name=viridis,
    enlargelimits=false,
    width=\columnwidth,
    title style={yshift=-0.2cm},
    label style={font=\footnotesize},      
    every tick label/.append style={font=\footnotesize},    
},
empty colorbar/.style={
    colorbar,
    colorbar style={
        nodes near coords={},    
        ytick=\empty,
   }
}
}

\usepackage{booktabs}
\usepackage{makecell}

\usepackage{caption}
\usepackage{enumitem}

\usepackage{tikz}
\usetikzlibrary{shapes.geometric, arrows.meta, positioning, shadows, fit}
\usepackage{forest}

\usepackage{arydshln}

\makeatletter
\def\adl@drawiv#1#2#3{%
        \hskip.5\tabcolsep
        \xleaders#3{#2.5\@tempdimb #1{1}#2.5\@tempdimb}%
                #2\z@ plus1fil minus1fil\relax
        \hskip.5\tabcolsep}
\newcommand{\cdashlinelr}[1]{%
  \noalign{\vskip\aboverulesep
           \global\let\@dashdrawstore\adl@draw
           \global\let\adl@draw\adl@drawiv}
  \cdashline{#1}
  \noalign{\global\let\adl@draw\@dashdrawstore
           \vskip\belowrulesep}}
\makeatother

\usepackage{listings}

\usepackage{array}

\definecolor{cvprblue}{rgb}{0.21,0.49,0.74}

\usepackage[pagebackref,breaklinks,colorlinks,allcolors=cvprblue]{hyperref}

\def\paperID{16338} %
\def\confName{CVPR}
\def\confYear{2025}
\newcommand{\parsection}[1]{\textbf{#1:}}
\title{ProHOC: Probabilistic Hierarchical Out-of-Distribution Classification via Multi-Depth Networks}

\author{Erik Wallin\textsuperscript{1,2}, Fredrik Kahl\textsuperscript{2}, Lars Hammarstrand\textsuperscript{2} \\
\textsuperscript{1}Saab AB, \textsuperscript{2}Chalmers University of Technology \\
{\tt\small \{walline,fredrik.kahl,lars.hammarstrand\}@chalmers.se}}

\begin{document}
\maketitle
\begin{abstract}
Out-of-distribution (OOD) detection in deep learning has traditionally been framed as a binary task, where samples are either classified as belonging to the known classes or marked as OOD, with little attention given to the semantic relationships between OOD samples and the in-distribution (ID) classes. We propose a framework for detecting and classifying OOD samples in a given class hierarchy. Specifically, we aim to predict OOD data to their correct internal nodes of the class hierarchy, whereas the known ID classes should be predicted as their corresponding leaf nodes. Our approach leverages the class hierarchy to create a probabilistic model and we implement this model by using networks trained for ID classification at multiple hierarchy depths. We conduct experiments on three datasets with predefined class hierarchies and show the effectiveness of our method. Our code is available at \url{https://github.com/walline/prohoc}.

\end{abstract}    
\section{Introduction}
\label{sec:intro}

Effectively handling out-of-distribution samples is important for deep-learning applications. Data outside the training domain often yield unpredictable results when fed through deep neural networks \cite{hendrycks2016baseline}, making it essential to account for previously unseen data when deploying these models in real-world settings to ensure robust performance and avoid unexpected outcomes. Previous literature \cite{zhang2023openood} has focused on binary out-of-distribution detection, predicting data either as ID (in-distribution) or OOD (out-of-distribution), without differentiating between OOD samples that are semantically close or far from ID. For example, with ID classes consisting of dog breeds, this binary paradigm would treat an image of an unknown dog the same as an image of an airplane, overlooking the varying degrees of semantic similarity these samples have to the known classes.

\begin{figure}[!t]
    \centering
    \hspace{-0.7cm}
    \input{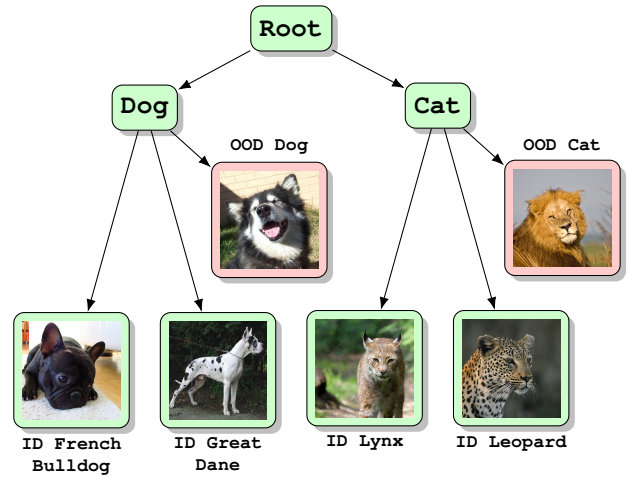} \vspace{-0.3cm}
    \caption{Out-of-distribution detection in class hierarchies. Instead of simply predicting the unseen dog and cat types as OOD, we aim to classify them as the high-level categories dog and cat.}
    \label{fig:intro}
\end{figure}

We explore a new setting of OOD detection, utilizing class hierarchies to predict semantically close OOD samples as their correct nodes in the class hierarchy. Hierarchical models for organizing and classifying objects are widespread, from foundational examples like Linnaeus' taxonomy in \emph{Systema Naturae} \cite{linnaeus1789systema} to modern structures like WordNet \cite{miller1995wordnet} for semantic relations. While the deep learning community has shown a growing interest in utilizing class hierarchies \cite{bertinetto2020making, jain2024test}, their application for OOD detection remains largely unexplored. However, such hierarchies can provide a foundation for enabling more informative OOD predictions. For instance, in the hierarchy shown in \cref{fig:intro}, the binary paradigm of OOD detection treats the unseen cat and dog as simply OOD. In contrast, with knowledge of the hierarchy, we can classify the unseen dog as the broader dog category (while recognizing it as distinct from previously seen breeds).

One of many motivating examples arises in the context of an autonomous driving system. A sensor model might be trained to classify objects such as bicycles, electric scooters, and mopeds (among other classes). However, many system components treat these objects similarly, grouping them as slow-moving vehicles. When encountering an obscure object not present in the training data, such as a unicycle or a penny-farthing, the ideal response would be to recognize it as a slow-moving vehicle, rather than misclassifying it as a known class or simply marking it as OOD, which would provide limited actionable information.

Our approach addresses this problem by leveraging the class hierarchy to factorize the probability distribution of predictions in the tree. This probability model includes conditional probabilities at each parent node, at which the possible predictions are either one of its child nodes or OOD at that specific node. Our key challenge lies in modeling these conditionals, as we assume no OOD data are available for training. To approximate these conditionals, we train classification networks at each depth of the hierarchy. We find that networks trained to classify classes higher in the hierarchy better recognize features associated with broad categories, making them better at predicting OOD samples to the correct high-level category. Similarly, the more fine-grained models may display larger uncertainty for these OOD samples because they contain low-level features that are not recognized by these models.

We explore several methods for leveraging these multi-depth networks to approximate the conditionals in our probabilistic framework based on standard OOD scores. We evaluate our proposed model, \emph{ProHOC},  on three datasets with predefined class hierarchies that we split into ID and OOD and show that our method outperforms previous attempts. Moreover, our framework has the benefit of introducing no new hyperparameters and uses standard methods for training the underlying multi-depth networks, making it straightforward to extend and build upon. We believe that our model can serve as a strong foundation for further research in this relatively unexplored area.

Our main contributions are as follows:
\begin{itemize}
    \item We formulate a probabilistic framework for classifying data, both ID and OOD, as nodes in a class hierarchy.
    \item We explore and enable the implementation of this framework by utilizing multi-depth networks to approximate the conditionals in the probabilistic model.
    \item We experimentally evaluate our framework and baselines on several datasets, showing the effectiveness of our method.
\end{itemize}

\section{Related work}
\label{sec:relatedwork}

\subsection{Hierarchy-aware ID classification}

Several works incorporate class hierarchies in training models for ID classification, focusing on minimizing the hierarchical distance between predictions and ground truth rather than only optimizing for classification accuracy \cite{bertinetto2020making, karthik2021no, garg2022learning, liang2023inducing, jain2024test}. This approach prioritizes models that make errors close to the correct class in the hierarchy, which often can be preferable to models that make more distant mistakes. This aspect of performance is typically disregarded in works on standard flat classification.

The existing approaches to this problem vary. In \cite{bertinetto2020making}, they propose adjustments to the standard cross-entropy loss to account for higher-level decisions in the hierarchy. The works \cite{garg2022learning, liang2023inducing} introduce hierarchy-aware feature spaces to obtain the desired model properties. In \cite{karthik2021no}, they suggest a post-hoc method that rescales flat prediction probabilities by hierarchical distances to produce predictions minimizing the expected hierarchical distance.

Most similar to our approach is \cite{jain2024test}, which proposes training separate coarse and fine-grained classifiers to improve both classification accuracy and hierarchical distance. However, their method is restricted to classifiers at two levels (one coarse and one fine), whereas we train classifiers at all levels of the hierarchy.

This overall line of research is similar to our setting in that we consider class hierarchies to minimize the hierarchical distance between ground truth and predictions. However, a key difference is that these works only consider ID data and therefore restrict the predictions to leaf nodes only. In contrast, we need a framework able to predict internal nodes as OOD predictions.

\subsection{Out-of-distribution detection}

Out-of-distribution is an active field of research \cite{scheirer2012toward, bendale2016towards, hendrycks2016baseline, lee2018simple, liang2017enhancing}. The goal is to detect whether a data sample belongs to the training distribution or not, motivated by the need to handle real-world, uncurated settings. The common approach in this field is to design a score based on neural network outputs, which takes high values for OOD data and low values for ID data (or vice versa). The score can be derived from, \eg, predicted distributions \cite{hendrycks2016baseline, ren2019likelihood, liu2023gen}, logits \cite{li2020energy, vaze2021open}, or feature representations \cite{lee2018simple, wang2022vim, ammar2024neco}.

Our setting introduces some new challenges compared to the standard binary paradigm of OOD detection. First, instead of only detecting OOD as a binary prediction, we aim to classify OOD samples as specific nodes within the class hierarchy. Second, we ultimately need to make hard decisions about which node to predict, so unnormalized scores that lack probabilistic interpretations or easily inferrable thresholds are of limited use. Lastly, many works on OOD detection use comparably simple problem settings, using one dataset as ID and an unrelated dataset as OOD \cite{lee2018simple, li2020energy}. In contrast, our setting naturally involves complex, fine-grained detection tasks, as OOD samples can appear at any depth within the class hierarchy.

As a final note, a few works exist that consider class hierarchies and OOD detection simultaneously \cite{vaze2021open, lang2024coarse,hogeweg2024cood}. In \cite{vaze2021open,lang2024coarse}, they use class hierarchies to construct OOD sets of varying difficulty by selecting the OOD sets from different depths of the class hierarchy. The work \cite{hogeweg2024cood} proposes a score that signals high OOD-ness if two classifiers predict distinct classes separated by a large hierarchical distance. However, these works still consider only binary OOD predictions.

\subsection{Hierarchical out-of-distribution detection}

To the best of our knowledge, the only work that considers detection and classification of OOD in class hierarchies similarly to us is \cite{linderman2023fine}. Their method involves separate model heads for each internal node in the hierarchy. For data that are descendants of a specific node, the corresponding head is trained to predict the correct child using a standard cross-entropy loss. For non-descendant data points, the head is trained to produce a uniform distribution.

There are key distinctions between \cite{linderman2023fine} and our work. First, our method introduces no additional hyperparameters, whereas \cite{linderman2023fine} requires balancing multiple loss terms during training. Second, their approach performs hierarchical inference in a top-down manner, with thresholds at each node to decide where to stop. These thresholds need to be assigned using ID data and it is not clear how to select these to ensure performance on both ID and OOD data while generalizing to multiple datasets. In contrast, our fully probabilistic approach provides a predictive distribution over all nodes in the hierarchy, eliminating the need for threshold-based top-down inference.

\section{A probabilistic hierarchy-framework} \label{sec:prob-framework}

Our approach for detecting and classifying OOD in class hierarchies is based on creating a probabilistic model from the class hierarchy. We assume that this hierarchy $\mathcal{H}$ is given and that it is structured as a \emph{directed rooted tree} \cite{bender2010lists} (\eg, the green nodes of \cref{fig:intro}). We denote the nodes of this tree as $\mathcal{C}$ with the leaf nodes being $\mathcal{C}^\text{id} \subset \mathcal{C}$. We assume a distribution of ID data with classes corresponding to leaves in this hierarchy: $p^\text{id}(x,y)$, with $y \in \mathcal{C}^\text{id}$. These data are observed during training. Furthermore, we have a distribution of OOD data, not observable during training, $p^\text{ood}(x,y)$, with $y \in \mathcal{C} \setminus \mathcal{C}^\text{id}$. In other words, the OOD data do not match any ID leaf classes but correspond to groups of ID classes higher in the hierarchy.

Our goal is to learn a model $f(x)$ that predicts samples from the balanced mix of ID and OOD, defined as $p^\text{mix}(x,y) = 0.5(p^\text{id}(x,y) + p^\text{ood}(x,y))$, to their correct nodes in $\mathcal{H}$. If misclassified, the predictions should be hierarchically close to the ground truth. Formally, we aim to minimize the expected hierarchical distance between the prediction and the ground truth:
\begin{equation} \label{eq:objective}
    \min \mathbb{E}_{p^\text{mix}(x,y)} \left[ \text{dist}_\mathcal{H}(f(x),y) \right],
\end{equation}
where $\text{dist}_\mathcal{H}(\cdot, \cdot)$ is the number of edges in the shortest path between two nodes in the undirected equivalent of $\mathcal{H}$.

To model $f(\cdot)$, we construct a probabilistic model from the class hierarchy $\mathcal{H}$. Specifically, we append a child node to each internal node of $\mathcal{H}$ to represent OOD predictions at that node (\eg, the red nodes of \cref{fig:intro}), creating the new tree $\mathcal{G}$. We denote the set of these OOD nodes as
\begin{equation}
    \mathcal{C}^\text{ood} = \left\{ \text{ood}(c) | c \in \mathcal{C} \setminus \mathcal{C}^\text{id} \right\}.
\end{equation}
These nodes represent a sample belonging to class $c$ but not to any of $c$'s known descendants. The model $\mathcal{G}$ describes the semantic classes of a sample as random binary variables at each node. We denote the probability of the set of nodes $\mathcal{C}'$ being active for a given sample $x$ as $p(\mathcal{C}'|x)$. We assume that all samples belong to a leaf node $c \in \mathcal{C}^\text{id} \cup \mathcal{C}^\text{ood}$ but that the leaf nodes are \emph{mutually exclusive}, \ie, a sample belongs to only one leaf node. Moreover, the internal nodes of $\mathcal{G}$ describe unions of these leaf nodes such that internal node $c$ is active if any of its descendant leaf nodes are active.  This implies $p(c|x) = p(\text{Anc}_\mathcal{G}^+(c)|x)$, where $\text{Anc}_\mathcal{G}^+(c)$ is the set of $c$ and all $c$'s ancestors. From this follows the conditional independences $p(c|\text{Anc}_\mathcal{G}^+(\text{Par}_\mathcal{G}(c))) = p(c|\text{Par}_\mathcal{G}(c))$ where $\text{Par}_\mathcal{G}(c)$ is the parent node of $c$.

We can now factorize the probability of paths in $\mathcal{G}$ using the chain rule of probability and the induced conditional independences as
\begin{equation} \label{eq:anc-joint2}
    p(c|x) = p(\text{Anc}_\mathcal{G}^+(c)|x) = \prod_{\mathclap{c' \in \text{Anc}_\mathcal{G}^+(c) \setminus R}} p(c'|\text{Par}_\mathcal{G}(c'), x),
\end{equation}
where the root node $R$ is excluded in the product because $p(R|x)=1$. For example, in the hierarchy of \cref{fig:intro}, the probability for \emph{Lynx} becomes $p(\text{Lynx}|\text{Cat},x)p(\text{Cat}|R,x)$ whereas the probability for an OOD Cat is $p(\text{ood}(\text{Cat})|\text{Cat},x)p(\text{Cat}|R,x)$.

By applying \eqref{eq:anc-joint2} with $c \in \mathcal{C}^\text{id} \cup C^\text{ood}$, we obtain the predictive distribution over all ID classes and OOD predictions at each internal node of $\mathcal{H}$. Note that for evaluation purposes, predictions of $\text{ood}(c)$ are mapped to the corresponding node $c$ in $\mathcal{H}$ to compute hierarchical distances.

We use this distribution to obtain our final prediction. A standard approach is to use the argmax of $p(c|x)$ with $c \in \mathcal{C}^\text{id} \cup \mathcal{C}^\text{ood}$. However, to better align with our objective in \eqref{eq:objective}, we instead utilize the uncertainties of the predictive distribution to minimize the expected hierarchical distance between the predicted node for sample $x$ and its ground truth:
\begin{equation} \label{eq:min-hdist}
    f(x) = \argmin_{c \in \mathcal{C}^\text{id} \cup \mathcal{C}^\text{ood}}  \mathbb{E}_{p(c'|x)} \left[ \text{dist}_\mathcal{H}(c,c')\right]
\end{equation}
where the expectation is obtained by
\begin{equation} 
    \mathbb{E}_{p(c'|x)} \left[ \text{dist}_\mathcal{H}(c,c') \right] = \sum_{\mathclap{c' \in \mathcal{C}^\text{id} \cup \mathcal{C}^\text{ood}}} \text{dist}_\mathcal{H}(c,c')p(c'|x).
\end{equation}
Note that $\text{ood}(c)$ is mapped to $c$ in $\text{dist}_\mathcal{H}(\cdot,\cdot)$. In \cref{sec:ablation-hdist}, we experimentally show the benefit of using \eqref{eq:min-hdist} compared to the standard argmax approach.

\section{Leveraging multi-depth networks}\label{sec:multi-depth}

The predictive distribution of \eqref{eq:anc-joint2} requires the conditional distributions at each internal node of the class hierarchy:
\begin{equation} \label{eq:conditionals}
    p(y|c,x), \quad y \in \text{Ch}_\mathcal{H}(c) \cup \{ \text{ood}(c) \}
\end{equation}
for $c \in \mathcal{C} \setminus \mathcal{C}^\text{id}$, where $\text{Ch}_\mathcal{H}(c)$ is the set of children of $c$ according to the original tree $\mathcal{H}$. However, as no OOD data are available during training, we cannot train a model to directly predict $p(\text{ood}(c)|c,x)$. Instead, as common for binary OOD detection \cite{lee2018simple, wang2022vim}, we resort to hand-crafted models for the OOD predictions.

Our idea for approaching these conditionals is to design classifiers from ID data that, given a sample belonging to $\text{ood}(c)$, confidently predict class $c$ while expressing uncertainty for the known child classes $\text{Ch}_\mathcal{H}(c)$. With such a model, the uncertainty of the sample's membership in $\text{Ch}_\mathcal{H}(c)$ can be used as a proxy for $p(\text{ood}(c)|c,x)$.

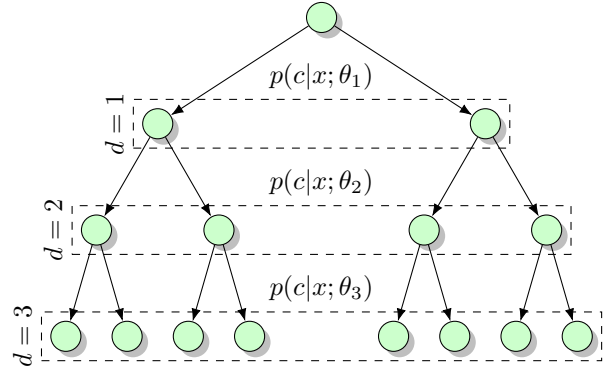
\begin{figure}[t]
    \centering
    \begin{forest}
    for tree={
        circle,
        draw,
        minimum size=4mm, %
        font=\scriptsize,
        edge={-Latex}, %
        l sep=1cm, %
        s sep=0.4cm, %
        fill=green!20,
        drop shadow
    },
    where n children=0{tier=word}{}    
    [,s sep=1.5cm
        [, name=A 
            [, name=A1
                [, name=A11]
                []
            ]
            [
                []
                []
            ]
        ]
        [, name=B
            [
                []
                []
            ]
            [, name=B2
                []
                [, name=B22]
            ]
        ]
    ]
    \node[draw,dashed,fit=(A) (B), label=above:{$p(c|x;\theta_1)$}, label={[rotate=90, anchor=south]left:{$d=1$}}] {};
    \node[draw,dashed,fit=(A1) (B2), label=above:{$p(c|x;\theta_2)$}, label={[rotate=90, anchor=south]left:{$d=2$}}] {};
    \node[draw,dashed,fit=(A11) (B22), label=above:{$p(c|x;\theta_3)$}, label={[rotate=90, anchor=south]left:{$d=3$}}] {};
\end{forest} \vspace{-0.3cm}
    \caption{For a hierarchy of depth 3, we train separate neural networks, parameterized by $\theta_1$, $\theta_2$, and $\theta_3$, to classify data into nodes within that depth level.}
    \label{fig:multi-depth}
\end{figure}

Following this reasoning, we propose training separate classification networks for the different depths of the hierarchy, with each network responsible for classifying data into nodes at a particular depth (see \cref{fig:multi-depth}). The intuition behind this approach is that the networks for high levels will emphasize the features associated with broad categories, which may not be helpful for more fine-grained classification. For instance, in the dogs-and-cats hierarchy of \cref{fig:intro}, the network trained for the level above the leaves, responsible for classifying \emph{dog} or \emph{cat}, may focus on features like ear and tail shape, useful for distinguishing between any dog and any cat. Conversely, the leaf-level network might focus on fur patterns or facial features, important for distinguishing specific breeds. Consequently, an unseen cat breed might lack the features the leaf model recognizes, causing it to show uncertainty. However, the high-level network could still identify the features common to all cats and confidently classify the unseen cat breed as a cat.

Specifically, we train these \emph{multi-depth networks} using an ID dataset of samples and labels
\begin{equation}
    \mathcal{S}^\text{id} = \left\{ (x_i, y_i) | y_i \in \mathcal{C}^\text{id} \right\}_{i=1}^{n_\text{id}}.
\end{equation}
To train high-level models, we map these leaf-level labels to their corresponding ancestors at specific depths. For this purpose, we define the function $\lambda(y,d)$ which maps the label $y$ to its corresponding ancestor at depth $d$, or if $\text{Depth}_\mathcal{H}(y) \leq d$, maps $y$ to itself (this handles the case of all leaves not having equal depth).
Consequently, we train the network for depth $d$, denoted $f_{\theta_d}$, using standard cross-entropy training, with the dataset
\begin{equation} \label{eq:multidepth-labels}
    \mathcal{S}_d^\text{id} = \left\{ (x_i, \lambda(y_i,d)) | y_i \in \mathcal{C}^\text{id} \right\}_{i=1}^{n_\text{id}},
\end{equation}
and obtain the multi-depth networks $f_{\theta_1}, \dots, f_{\theta_D}$ with the parameters $\theta_1, \dots, \theta_D$, where $D=\max_{c \in \mathcal{C}} \text{Depth}_\mathcal{H}(c)$.

In \cref{sec:multi-depth-results}, we present results showing that our multi-depth networks effectively classify OOD data.

\subsection{Modeling the conditionals of the hierarchy} \label{sec:conditionals}

To model the conditionals $p(y|c,x)$ based on the multi-depth networks, $f_{\theta_1}, \dots, f_{\theta_D}$, we are looking for ways to quantify the uncertainty of $x$ belonging to $\text{Ch}_\mathcal{H}(c)$ and use this to estimate $p(\text{ood}(c)|c,x)$. There are a variety of ways one could approach this. We have evaluated several alternatives based on standard OOD scores in \cref{sec:ablation-scores}. An intuitive and simple approach is to use
\begin{equation}
    p(y|c,x) = p(y|x;\theta_d) \quad \text{for} \quad y \in \text{Ch}_\mathcal{H}(c)
\end{equation}
and
\begin{equation} \label{eq:sum-score}
    p(\text{ood}(c)|c,x) = 1 - \sum_{\mathclap{y \in \text{Ch}_\mathcal{H}(c)}} p(y|x;\theta_d),
\end{equation}
where $d = \text{Depth}_\mathcal{H}(c)+1$. This formulation uses the probability of a sample belonging to a class outside the children of $c$ as a proxy for the OOD probability while keeping the conditional probabilities for the children as predicted by $p(y|x;\theta_d)$. However, we find that this approach alone tends to assign too low probabilities for OOD and fails to account for the uncertainty among the child nodes. To address this, we also consider the entropy of the normalized distribution over the children:
\begin{equation} \label{eq:entropy-score}
    H(c) = - \sum_{\mathclap{c' \in \text{Ch}_\mathcal{H}(c)}} \tilde{p}(c'|x;\theta_d) \log \tilde{p}(c'|x;\theta_d),
\end{equation}
where
\begin{equation}
    \tilde{p}(c'|x;\theta_d) = \frac{p(c'|x;\theta_d)}{\sum_{\tilde{c} \in \text{Ch}_\mathcal{H}(c)} p(\tilde{c}|x;\theta_d)}.
\end{equation}
High entropies indicate uncertainty among the children, \ie, when multiple children are assigned similar probabilities.

Since the entropy and the complementary probability \eqref{eq:sum-score} are independent and account for different aspects of uncertainty, we combine these in a sum as
\begin{equation} \label{eq:entropysum-score}
    s(c) = H(c) + 1 - \sum_{\mathclap{y \in \text{Ch}_\mathcal{H}(c)}} p(y|x;\theta_d).
\end{equation}
However, this score no longer forms a probability distribution when combined with $p(y|x;\theta_d)$ for $y \in \text{Ch}_\mathcal{H}(c)$. To resolve this, we normalize to obtain the valid distribution as
\begin{equation} \label{eq:renorm-id}
    p(y|c,x) = \frac{p(y|x;\theta_d)}{s(c) + \sum_{y' \in \text{Ch}_\mathcal{H}(c)} p(y'|x;\theta_d)}
\end{equation}
for $y \in \text{Ch}_\mathcal{H}(c)$ and
\begin{equation} \label{eq:renorm-ood}
    p(\text{ood}(c)|c,x) = \frac{s(c)}{s(c) + \sum_{y' \in \text{Ch}_\mathcal{H}(c)} p(y'|x;\theta_d)},
\end{equation}
which gives us the final model for the conditional distributions in our hierarchical framework.

\section{Experiments and results}
\label{sec:experiments}

\subsection{Datasets} \label{sec:datasets}

There are no established datasets for OOD detection in class hierarchies. We construct our benchmarks using three datasets with predefined hierarchies commonly used in work for hierarchy-aware ID classification. We generate ID and OOD subsets by selecting OOD nodes from various depths within these hierarchies. Our experiments are conducted on the following datasets:

\parsection{FGVC-Aircraft \cite{fgvcaircraft}} 
This dataset contains aircraft images with a three-level hierarchy: at the highest level are manufacturers (\eg, Boeing), the second level groups aircraft by family (\eg, Boeing 747), and the most specific level contains variants (\eg, Boeing 747-200). FGVC-Aircraft has the most shallow hierarchy and the fewest classes among the considered datasets, but it poses challenging classification tasks, such as distinguishing between separate Boeing 747 types.

\parsection{iNaturalist19 \cite{van2018inaturalist}} 
A dataset of biological species organized in a hierarchy according to taxonomic ranks such as \emph{kingdom}, \emph{phylum}, and \emph{genus}. It includes a large number of classes with fine-grained relationships and has a deeper hierarchy than FGVC-Aircraft.

\parsection{SimpleHierImageNet}
This dataset is based on tieredImageNet \cite{ren18fewshotssl}, which is a subset of ImageNet using a hierarchy based on the WordNet graph \cite{miller1995wordnet}. However, the hierarchy of tieredImageNet has many nodes representing conceptual rather than visual similarities. For example, it has the high-level class \emph{timepiece} with the subclasses \emph{analog clock} and \emph{digital clock}, or the leaf class \emph{laptop computer} under the high-level class \emph{portable computer}, while \emph{computer keyboard} is nine edges away under \emph{electronic equipment}. To address this, we introduce SimpleHierImageNet by reorganizing parts of the tieredImageNet hierarchy to reflect visual similarities. Despite these adjustments, SimpleHierImageNet remains a challenging problem with a wide range of categories and a deep hierarchy with many internal nodes. More details on SimpleHierImageNet are available in the supplementary material.

The considered datasets lack established ID/OOD splits. Therefore, we construct the OOD set by selecting nodes from various depths of the hierarchy. For each selected OOD node, we remove it and all its descendants from the hierarchy, leaving the remaining nodes to form our ID hierarchy $\mathcal{H}$. After defining the ID hierarchy $\mathcal{H}$ and the set of OOD classes, we map the labels of the OOD data (which are leaf nodes in the full original hierarchy) to their closest ancestor present in the ID hierarchy $\mathcal{H}$. These mapped labels are then used in the experimental evaluation.

Details of the datasets, after defining the ID and OOD sets are described in \cref{tab:dataset-details}. Additional information on the datasets and the selection of OOD sets is available in the supplementary material.

\begin{table}[]
    \centering
    \footnotesize
    \setlength{\tabcolsep}{5pt}
    \caption{Dataset details (after being split into ID and OOD).} \vspace{-0.3cm}
    \label{tab:dataset-details}    
    \begin{tabular}{lcccc}
    \toprule
     & \makecell{Max\\depth} & \makecell{Nr. ID\\leaves} & \makecell{Nr. internal\\nodes} & \makecell{Nr. OOD\\classes} \\ \midrule
    FGVC-Aircraft & 3 & 80 & 28 & 20 \\
    SimpleHierImageNet & 11 & 518 & 63 & 80 \\
    iNaturalist19 & 6 & 721 & 77 & 289 \\
    \bottomrule
\end{tabular}

\end{table}

\subsection{Evaluating multi-depth networks for OOD} \label{sec:multi-depth-results}

To evaluate our hypothesis that multi-depth networks are effective for OOD classification, we compare the accuracy obtained from our high-level models $f_{\theta_1},\dots,f_{\theta_{D-1}}$ to the accuracy obtained by marginalizing the leaf-level model $f_{\theta_D}$, when evaluated on OOD data.

Specifically, we have a dataset of OOD data associated with different depths of the hierarchy:
\begin{equation}
    \mathcal{S}^\text{ood} = \left\{ (x_i, y_i) | y_i \in \mathcal{C} \setminus \mathcal{C}^\text{id} \right\}_{i=1}^{n_\text{ood}}.
\end{equation}
We define subsets of $\mathcal{S}^\text{ood}$ associated with specific depths as
\begin{equation}
    \mathcal{S}^\text{ood}_d = \left\{ (x_i, y_i) \in \mathcal{S}^\text{ood} | \text{Depth}_\mathcal{H}(y_i) = d \right\},
\end{equation}
for $d \in \{1,\dots,D-1\}$. Now we can compute the classification accuracies obtained by the network $f_{\theta_d}$ on the dataset $\mathcal{S}_d^\text{ood}$. We compare this with the predictions obtained by marginalizing the predictions of the ID leaf model $f_{\theta_D}$ as
\begin{equation}
    p_d^\text{margin}(c|x) = \sum_{\mathclap{c' \in \text{Leaves}_\mathcal{H}(c)}} p(c' | x; \theta_D) \quad \text{for}~c \in \mathcal{C}^d,
\end{equation}
where $\text{Leaves}_\mathcal{H}(c)$ are the descendant leaves of $c$.

\Cref{tab:leaf-vs-coarse} shows the results of this comparison across our three considered datasets. The multi-depth model obtains higher accuracies than the marginalized model for all datasets, indicating that the high-level models are better at classifying OOD samples as the correct class. Moreover, this implies that while a high-level model correctly predicts an OOD sample as class $c$, the lower-level model leans towards predicting classes not in $\text{Ch}_\mathcal{H}(c)$, which is the behavior we are looking for.

\begin{table}[]
    \centering
    \caption{Classification accuracies on OOD data at ground-truth depths. We compare models trained with high-level labels with models trained using the leaf labels.} \vspace{-0.3cm}
    \label{tab:leaf-vs-coarse}    
    \footnotesize
    \begin{tabular}{cccc} 
    \toprule
    Dataset & \multicolumn{2}{c}{Accuracy (\%) $\uparrow$} & $\Delta$ $\uparrow$ \\ \cmidrule{1-1} \cmidrule(l){2-3} \cmidrule(l){4-4} 
    & Multi-depth & Marginalized & \\
    & $p(c|x; \theta_d)$ & $p_d^\text{margin}(c|x)$ & \\ \midrule
    SimpleHierImageNet & 70.1 & 68.5 & +1.6 \\
    iNaturalist19 & 76.5 & 73.6 & +2.9 \\
    FGVC-Aircraft & 67.3 & 50.3 & +17.0 \\
    \bottomrule
\end{tabular}
\end{table}

\subsection{Evaluation metrics} \label{sec:metrics}

How to evaluate a framework for OOD detection in class hierarchies is not established. In alignment with works on ID classification with class hierarchies \cite{karthik2021no, liang2023inducing}, we focus primarily on hierarchical distances between predictions and ground truth. However, given that our OOD nodes are selected at varying depths in the hierarchy, the resulting OOD sets have highly imbalanced class distributions. Therefore we propose to use a balanced mean hierarchical distance (BMHD), defined as
\begin{equation} \label{eq:bmhd}
    \text{BMHD}(\mathcal{C}') = \frac{1}{|\mathcal{C}'|} \sum_{c \in \mathcal{C}'} \frac{1}{n_c} \sum_{i=1}^{n_c} \text{dist}_\mathcal{H}(f(x_i^c),c),
\end{equation}
where $\mathcal{C}'$ is the set of classes being considered, $n_c$ is the number of samples in the test set with ground truth label $c$, $x_i^c$ is the $i$-th sample out of those, and $|\mathcal{C}'|$ is the cardinality of $\mathcal{C}'$. In $\text{BMHD}(\mathcal{C}')$, the mean distances for nodes in $\mathcal{C}'$ are weighted equally, regardless of how many samples are associated with a specific node.

Furthermore, we want to evaluate the performance on a combined set of ID and OOD data. To this end, we define the mixed BMHD as
\begin{equation}
    \text{MixBMHD} = 0.5 \left( \text{BMHD}_\text{id} + \text{BMHD}_\text{ood} \right),
\end{equation}
where $\text{BMHD}_\text{id}$ and $\text{BMHD}_\text{ood}$ are evaluated with \eqref{eq:bmhd} on the set of ID and OOD classes with test data available, respectively. Similarly, we evaluate classification accuracies as the proportion of predictions that match the ground truth labels. Because of the class imbalance, we consider a balanced accuracy score \cite{mosley2013balanced}:
\begin{equation}
    \text{MixBAcc} = 0.5 \left( \text{BAcc}_\text{id} + \text{BAcc}_\text{ood} \right),
\end{equation}
where $\text{BAcc}_\text{id}$ and $\text{BAcc}_\text{ood}$ denote the balanced accuracies for ID and OOD classes, respectively.

\subsection{Results} \label{sec:main-results}

We compare our framework to the following baselines:

\parsection{Depth oracle} We make predictions using the ground-truth depth with the corresponding $p(c|x;\theta_d)$ for all data, $\ie$, $p(c|x;\theta_D)$ for ID data and $p(c|x;\theta_d), d<D$ for OOD data. Given the accuracies of the multi-depth networks, we consider this an upper performance bound.

\parsection{Leaf model} We predict all data using $p(c|x;\theta_D)$, \ie, all data are predicted as leaves, including OOD.

\parsection{HSC} We train and evaluate the model proposed in \cite{linderman2023fine} using the authors' code. We present results using the \emph{synset-based stopping-criterion} because that gives the best results. We set thresholds for stopping criteria based on TPR rates to minimize MixBMHD on the test sets. However, in real-world settings, we cannot optimize these thresholds based on test performance.

Our framework, denoted \textbf{ProHOC} (for probabilistic hierarchical OOD classification), is evaluated using the two different models for the conditionals described in \cref{sec:conditionals}. The first, denoted CompProb, uses the complementary probabilities to model the OOD probabilities, as defined in \eqref{eq:sum-score}. The second, EntCompProb, uses the sum of the entropy and the complementary probability, as defined in \eqref{eq:entropysum-score}.

\begin{table*}[!t]
    \centering
    \caption{Mean hierarchical distances and accuracies on ID, OOD, and the mix of ID and OOD for ProHOC and baselines. The oracle model serves as an upper performance bound. Excluding the oracle model, the best results are \textbf{boldfaced}. Evaluations use test sets.} \vspace{-0.3cm}
    \label{tab:main-results}
    \footnotesize
    \begin{tabular}{l >{\color{gray}}c >{\color{gray}}c c >{\color{gray}}c >{\color{gray}}c c} %
 \toprule
 & $\text{BAcc}_\text{id}$ $\uparrow$ & $\text{BAcc}_\text{ood}$ $\uparrow$ & MixBAcc $\uparrow$ & $\text{BMHD}_\text{id}$ $\downarrow$ & $\text{BMHD}_\text{ood}$ $\downarrow$ & MixBMHD $\downarrow$ \\ 
 \midrule
 & \multicolumn{6}{c}{\textsc{SimpleHierImageNet}} \\ \midrule
 Depth oracle & 79.7 & 72.5 & 76.1 & 0.82 & 1.05 & 0.93 \\ \cdashlinelr{1-7}
 Leaf model & 79.7 & 0.0 & 39.8 & 0.82 & 2.12 & 1.47 \\
 HSC \cite{linderman2023fine} & 76.8 & 7.8 & 42.3 & 0.77 & 1.78 & 1.28 \\
 ProHOC (CompProb) & 67.8 & 19.2 & 43.5 & 0.92 & 1.61 & 1.27 \\
 ProHOC (EntCompProb) & 62.5 & 30.3 & \textbf{46.4} & 0.96 & 1.45 & \textbf{1.21} \\ \midrule
 & \multicolumn{6}{c}{\textsc{iNaturalist19}} \\ \midrule
 Depth oracle & 72.4 & 75.9 & 74.2 & 0.85 & 0.82 & 0.83 \\ \cdashlinelr{1-7}
 Leaf model & 72.4 & 0.0 & 36.2 & 0.85 & 2.23 & 1.54 \\
 HSC \cite{linderman2023fine} & 68.1 & 8.4 & 38.2 & 0.80 & 1.76 & 1.28 \\
 ProHOC (CompProb) & 66.1 & 18.0 & 42.0 & 0.77 & 1.34 & 1.06 \\
 ProHOC (EntCompProb) & 57.7 & 35.6 & \textbf{46.7} & 0.78 & 1.10 & \textbf{0.94} \\ \midrule
 & \multicolumn{6}{c}{\textsc{FGVC-Aircraft}} \\ \midrule
 Depth oracle & 84.7 & 67.6 & 76.1 & 0.49 & 0.67 & 0.58 \\ \cdashlinelr{1-7}
 Leaf model & 84.7 & 0.0 & 42.3 & 0.49 & 2.00 & 1.25 \\
 HSC \cite{linderman2023fine} & 77.9 & 14.4 & 46.1 & 0.40 & 1.48 & 0.94 \\ 
 ProHOC (CompProb) & 80.1 & 17.1 & 48.6 & 0.41 & 1.25 & 0.83 \\
 ProHOC (EntCompProb) & 78.0 & 22.7 & \textbf{50.3} & 0.41 & 1.21 & \textbf{0.81} \\
 \bottomrule
 \end{tabular}
\end{table*}

The results are presented in \cref{tab:main-results}. Excluding the oracle model, ProHOC with EntCompProb yields the best MixBMHD and MixBAcc across all three datasets, with CompProb consistently ranking second. Looking at the numbers for ID and OOD separately, we note the trade-off between ID and OOD performance, where the CompProb model favors ID performance by making deeper predictions. In contrast, the EntCompProb model has a more balanced performance across ID and OOD.

ProHOC outperforms HSC \cite{linderman2023fine}, the existing method for OOD detection in hierarchies. This is despite optimizing the inference thresholds in HSC using OOD test data. We can see that HSC tends to make overly deep predictions, keeping an ID performance close to the leaf model but showing poor OOD performance. More detailed analyses of the results are available in the supplementary material.

\subsection{Evaluating out-of-hierarchy data}

\Cref{sec:main-results} focuses on evaluating within-hierarchy OOD data, as these samples are the most challenging and distinguish our hierarchical approach from the binary OOD setting. However, ProHOC supports out-of-hierarchy OOD by classifying such samples as OOD at the root node. The OOD probability at the root is computed via \eqref{eq:renorm-ood} with $s(\text{root})$ being a positive score from any suitable binary OOD method. For simplicity, we use the entropy of the deepest network for $s(\text{root})$ which is a common baseline for binary OOD detection \cite{ren2019likelihood} that aligns well with our EntCompProb model. Table \ref{tab:out-of-hierarchy} shows ProHOC for FGVC-Aircraft with root predictions on three out-of-hierarchy datasets. MixBAcc is the (balanced) accuracy on the mix of within-hierarchy ID and OOD. Acc for out-of-hierarchy datasets is the percentage of samples correctly classified as root. We get high accuracies on out-of-hierarchy OOD while keeping most within-hierarchy performance.

\begin{table}[]
    \centering
    \footnotesize
    \setlength{\tabcolsep}{2pt}
    \caption{Evaluating ProHOC on out-of-hierarchy data.} \vspace{-0.3cm}
    \label{tab:out-of-hierarchy}    
    \begin{tabular}{l c c c c}
\toprule
& MixBAcc $\uparrow$ & Acc $\uparrow$ & Acc $\uparrow$ & Acc $\uparrow$ \\ 
& (FGVC-Air.) & Office31 \cite{saenko2010adapting} & iNat19 & PACS \cite{li2017deeper} \\ \midrule
ProHOC w/ root & 47.1 & 78.2 & 83.9 & 94.4 \\
ProHOC w/o root & 50.3 & - & - & - \\ \bottomrule
\end{tabular}
\end{table}

\subsection{Training details}

We train our multi-depth networks with standard supervised NLL training using the remapped labels from \eqref{eq:multidepth-labels}. We use the ResNet50 architecture \cite{he2016deep} for all experiments. For FGVC-Aircraft and iNaturalist19, the network is initialized with pre-trained weights from ImageNet, whereas for SimpleHierImageNet, we train the network from scratch. We train for 90 epochs on FGVC-Aircraft and iNaturalist19, and 250 epochs on SimpleHierImageNet. We use stochastic gradient descent with an initial learning rate of 0.05, decayed to zero by the end of training using a cosine schedule. A batch size of 128 is used throughout. During training, images are randomly cropped and resized to $224 \times 224$.
\section{Limitations}

This work focuses on establishing the key building blocks of our framework for hierarchical OOD detection, with little attention given to optimizing the performance of the underlying multi-depth networks. Further work could explore stronger architectures and advanced training techniques to improve the performance of ProHOC. Finally, a fundamental limitation of our problem formulation is the assumption that the class hierarchy reflects observable visual similarities. If ID and OOD siblings do not share common visual features, we can not expect this type of framework to accurately predict OOD.
\section{Ablation studies}

\subsection{Evaluating different scores for the conditionals} \label{sec:ablation-scores}

In \cref{sec:conditionals} we described how we use the multi-depth networks to model the conditionals $p(y|c,x)$ for $y \in \text{Ch}_\mathcal{H}(c) \cup \{\text{ood(c)}\}$. The primary challenge in modeling $p(y|c,x)$ lies in assigning the OOD probability. Here, we explore several approaches to modeling this OOD probability, or an unnormalized OOD score, based on the predictions from our multi-depth networks. We consider the following models:

\parsection{CompProb}
The complementary probability, as defined in \eqref{eq:sum-score}. It uses the sum of probabilities for classes outside of $\text{Ch}_\mathcal{H}(c)$ as a proxy for the OOD probability.

\parsection{Entropy}
As defined in \eqref{eq:entropy-score}, indicates the uncertainty of predictions among the children. The entropy is widely used as an OOD score in the literature \cite{ren2019likelihood}.

\parsection{MaxProb}
The (negated) maximum probability among the children, defined as
\begin{equation}
    s^\text{MaxProb}(c) = 1 - \max_{y \in \text{Ch}_\mathcal{H}(c)} p(y|x;\theta_d)
\end{equation}
with $d = \text{Depth}_\mathcal{H}(c) + 1$. This alternative is similar to the widely used MSP score \cite{hendrycks2016baseline} for binary OOD detection.
    
\parsection{EntCompProb}
Since the entropy and the complementary probability are independent and signal different aspects of uncertainty, we find it intuitive to combine scores in a sum, as defined in \eqref{eq:entropysum-score}.

\parsection{CompLogits}
We sum the unnormalized logits for classes outside $\text{Ch}_\mathcal{H}(c)$ and use the sum in the softmax function alongside the logits for the child nodes. This method resembles CompProb, although it differs in the scaling of the OOD probability due to summing before applying the exponential function in the softmax.

The methods denoted CompProb and CompLogits form valid probability distributions directly by design. For the other alternatives, we obtain the conditional distribution by normalization following \eqref{eq:renorm-id} and \eqref{eq:renorm-ood}.

A well-performing model for the conditionals should display certain qualities. The probability for OOD in the conditional distribution should be the maximum when the sample is OOD. Moreover, if the network misclassifies an ID sample (predicting it as an incorrect ID child), we want our model to predict the OOD node instead of an incorrect ID prediction, which would increase the hierarchical distance from the ground truth. To assess these aspects of the performance, we use several metrics that are evaluated locally for a specific internal node:

\parsection{F1} The F1 score for binary classification where OOD is considered the positive label. A prediction is OOD if $p(\text{ood}(c)|c,x)$ is the largest element of $p(y|c,x)$, otherwise ID. We also report the corresponding FPR and TPR.

\parsection{Purity} The fraction of predicted ID samples, predicted as the correct child node.
    
\parsection{Dirty F1} Our variant of the F1 score where ID data that are misclassified as an incorrect child are labeled as OOD when computing the F1 score.

We evaluate these metrics for all parent nodes $c$ for which OOD data with ground truth at $c$ are available. As ID data, we use all data that belong to descendants of $c$. We report the mean metrics over all evaluated nodes.%

\Cref{tab:score-results} shows results from FGVC-Aircraft. EntCompProb achieves the highest F1 score. It has the largest FPR but this is compensated by its high purity, indicating that many false positives would have been classified as incorrect ID children if they were predicted as ID. For Dirty F1, EntCompProb again outperforms other methods, by an even larger margin. This suggests that the entropy-based uncertainty and the complementary probability interact effectively to create a model with the desired properties. In contrast, CompLogits performs poorly due to being underconfident on OOD (a low TPR). This implies that summation at the logit level leads to smaller OOD probabilities than summation at the probability level (as with CompProb).

\subsection{Minimizing the expected hierarchical distance} \label{sec:ablation-hdist}

ProHOC provides a predictive distribution over all nodes in the hierarchy, allowing us to incorporate uncertainty into our final prediction. Given our objective to produce predictions that are hierarchically close to the ground truth \eqref{eq:objective}, we can use the predicted distribution $p(c|x)$ to minimize the expected hierarchical distance of the prediction
\begin{equation} \label{eq:ablation-minhdist}
    f(x) = \argmin_c \mathbb{E}_{p(c'|x)} \left[ \text{dist}_\mathcal{H}(c,c') \right],
\end{equation}
as presented in \cref{sec:prob-framework}. This is similar to the post-hoc correction proposed in \cite{karthik2021no}, although they restrict the predictions to leaves. We compute the expected hierarchical distance over all nodes in the hierarchy.

We compare the predictions from \eqref{eq:ablation-minhdist} with the standard $\argmax_c p(c|x)$ in \cref{tab:min-expected-hdist}, using ProHOC with both CompProb and EntCompProb. As expected, \eqref{eq:ablation-minhdist} achieves lower hierarchical distances since it optimizes for that specifically. However, we also see improvements in accuracy as an additional benefit of using \eqref{eq:ablation-minhdist}.

\begin{table}[t]
    \centering
    \caption{Analyzing the performance of different methods for modeling $p(\text{ood}(c)|c,x)$. Results are from FGVC-Aircraft.} \vspace{-0.3cm}
    \label{tab:score-results}    
    \footnotesize
    \setlength{\tabcolsep}{4pt}
    \begin{tabular}{lrrrrrrr}
\toprule
             &    F1 $\uparrow$ &   FPR $\downarrow$ &   TPR $\uparrow$ &   Purity $\uparrow$ &   Dirty F1 $\uparrow$ \\
\midrule
 CompProb        & 0.47 & 0.09 & 0.42 &    0.88 &     0.49 \\
 Entropy     & 0.51 & 0.12 & 0.45 &    0.90 &     0.55 \\
 MaxProb        & 0.50 & 0.10 & 0.45 &    0.89 &     0.52 \\
 EntCompProb  & 0.56 & 0.15 & 0.52 &    0.90 &     0.59 \\
 CompLogits    & 0.17 & 0.01 & 0.14 &    0.87 &     0.18 \\
\bottomrule
\end{tabular}
    \vspace{-0.1cm}
\end{table}

\begin{table}[]
    \centering
    \caption{Comparing predictions from minimizing the expected hierarchical distance to predictions using the most probable class.} \vspace{-0.3cm}
    \label{tab:min-expected-hdist}    
    \footnotesize
    \setlength{\tabcolsep}{1pt}
    \begin{tabular}{lcccc}
\toprule
& \multicolumn{2}{c}{$\argmin \mathbb{E} [ \text{dist}_\mathcal{H} ]$} & \multicolumn{2}{c}{$\argmax p(c|x)$} \\ \cmidrule(l){2-3} \cmidrule(l){4-5}
& MixBAcc $\uparrow$ & MixBMHD $\downarrow$ & MixBAcc $\uparrow$ & MixBMHD $\downarrow$ \\ \midrule
\multicolumn{5}{c}{\textsc{SimpleHierImageNet}} \\ \midrule
EntCompProb & 46.4 & 1.21 & 45.8 & 1.25 \\
CompProb & 43.5 & 1.27 & 39.7 & 1.37 \\ \midrule
\multicolumn{5}{c}{\textsc{iNaturalist19}} \\ \midrule
EntCompProb & 46.7 & 0.94 & 45.6 & 0.98 \\
CompProb & 42.0 & 1.06 & 38.4 & 1.17 \\ \midrule
\multicolumn{5}{c}{\textsc{FGVC-Aircraft}} \\ \midrule
EntCompProb & 50.3 & 0.81 & 50.2 & 0.88 \\
CompProb & 48.6 & 0.83 & 47.4 & 0.90 \\ \bottomrule
\end{tabular}
    \vspace{-0.4cm}
\end{table}
\section{Conclusion}

The deep-learning community has made substantial progress in the binary setting of OOD detection. In this paper, we raise the bar for OOD detection by aiming for more informative predictions: predicting OOD as internal nodes in a class hierarchy. Our framework shows promising results. We also believe its simplicity in introducing no new hyperparameters and using underlying networks trained with standard methods makes it accessible and adaptable for further development. We hope this work will inspire continued exploration of OOD classification in class hierarchies and that ProHOC can serve as a foundation for even more effective methods. \nocite{oquab2023dinov2}
\section*{Acknowledgement}
\label{sec:acknowledgement}

This work was supported by Saab AB and Wallenberg AI, Autonomous Systems and Software Program (WASP) funded by the Knut and Alice Wallenberg Foundation. The computations were enabled by resources provided by the National Academic Infrastructure for Supercomputing in Sweden (NAISS), partially funded by the Swedish Research Council through grant agreement no. 2022-06725.
{
    \small
    \bibliographystyle{ieeenat_fullname}
    \bibliography{main}
}

\clearpage
\setcounter{page}{1}
\maketitlesupplementary

\section{Distributions of hierarchical distances}

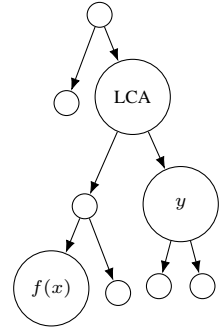
\begin{figure}[!t]
    \centering
    \begin{forest}
    for tree={
        circle,
        draw,
        font=\scriptsize,
        edge={-Latex}, 
    },   
    [
        []
        [LCA, minimum width=10mm,
            [
                [$f(x)$, minimum width=10mm,]
                []
            ]
            [$y$, minimum width=10mm,
                []
                []
            ]
        ]
    ]
\end{forest}
    \caption{Prediction example: $\text{dist}_\mathcal{H}(\text{LCA}, f(x)) = 2$, $\text{dist}_\mathcal{H}(\text{LCA}, y) = 1$.}
    \label{fig:lca-illustration-1}
\end{figure}

\begin{figure}[!t]
    \centering
    \begin{forest}
    for tree={
        circle,
        draw,
        align=center,
        font=\scriptsize,
        edge={-Latex}, 
    },   
    [
        []
        [LCA\\$f(x)$, minimum width=10mm,
            [
                []
                []
            ]
            [$y$, minimum width=10mm,
                []
                []
            ]
        ]
    ]
\end{forest}
    \caption{Prediction example: $\text{dist}_\mathcal{H}(\text{LCA}, f(x)) = 0$, $\text{dist}_\mathcal{H}(\text{LCA}, y) = 1$. This represents a \emph{pure underprediction}.}
    \label{fig:lca-illustration-3}
\end{figure}
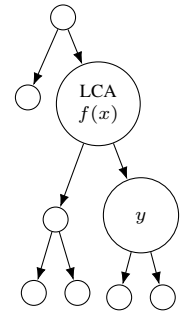

To analyze the performance of ProHOC in more detail, we compute histograms of hierarchical distances between the ground truth and the predictions. We compute these histograms for ProHOC with EntCompProb as the OOD model. For a more detailed evaluation of the hierarchical distances, we decompose these into an overprediction distance and an underprediction distance as
\begin{equation}
\begin{split}
    \text{dist}_\mathcal{H}(y, f(x)) = &\text{dist}_\mathcal{H}(\text{LCA}(y, f(x)), f(x)) \\ 
    + &\text{dist}_\mathcal{H}(\text{LCA}(y, f(x)), y)
    \end{split}
\end{equation}
where $y$ is the ground-truth node, $f(x)$ is the predicted node and $\text{LCA}(y, f(x))$ is the lowest common ancestor of $y$ and $f(x)$, \ie, the deepest node that has both $y$ and $f(x)$ as descendants (where descendants includes itself). We will use $\text{LCA} = \text{LCA}(y, f(x))$ for brevity. With this decomposition, we get the following error cases:
\begin{itemize}
    \item The prediction is deeper than the LCA: $\text{dist}_\mathcal{H}(\text{LCA}, f(x)) > 0$. 
    \item The ground truth is deeper than the LCA: $\text{dist}_\mathcal{H}(\text{LCA}, y) > 0$.
    \item The prediction is a descendant of the ground truth: $\text{dist}_\mathcal{H}(\text{LCA}, f(x)) > 0$ and $\text{dist}_\mathcal{H}(\text{LCA}, y) = 0$. We denote this case \emph{pure overprediction}.  
    \item The prediction is an ancestor of the ground truth: $\text{dist}_\mathcal{H}(\text{LCA}, f(x)) = 0$ and $\text{dist}_\mathcal{H}(\text{LCA}, y) > 0$. We denote this case \emph{pure underprediction}.
\end{itemize}
\Cref{fig:lca-illustration-1,fig:lca-illustration-2,fig:lca-illustration-3} illustrates these concepts.

The decomposed hierarchical distances are shown in \cref{fig:distances-fgvc-aircraft,fig:distances-inat,fig:distances-simple-hier-imagenet} where each sample in the respective test sets contributes to a histogram entry.

For OOD data, we observe both over- and underpredictions. Notably, pure overprediction distances of 1 are frequent across all three datasets. In contrast, ID data shows a clear trend of pure underprediction, with many samples being predicted as ancestors to the ground truth. As discussed in \cref{sec:main-results}, ProHOC with EntCompProb generally demonstrates lower ID performance compared to the other models. However, these histograms reveal that the low ID performance is primarily due to predicting ancestors to the ground truth, a behavior that may be acceptable in some applications.

\begin{figure}[!t]
    \centering
    \begin{forest}
    for tree={
        circle,
        draw,
        align=center,
        font=\scriptsize,
        edge={-Latex}, 
    },   
    [
        []
        [{LCA\\$y$}, minimum width=10mm,
            [
                [$f(x)$, minimum width=10mm,]
                []
            ]
            [
                []
                []
            ]
        ]
    ]
\end{forest}
    \caption{Prediction example: $\text{dist}_\mathcal{H}(\text{LCA}, f(x)) = 2$, $\text{dist}_\mathcal{H}(\text{LCA}, y) = 0$. This represents a \emph{pure overprediction}.}
    \label{fig:lca-illustration-2}
\end{figure}
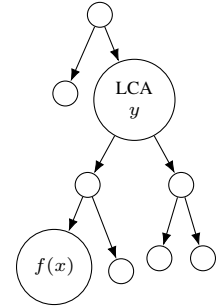

\clearpage

\begin{figure*}[!h]
    \centering
    \begin{tikzpicture}

\begin{groupplot}[
    group style={
        group size=2 by 1,
        vertical sep=1.0cm,
        },
    hdist base,
    width=0.7\columnwidth,
    height=0.7\columnwidth,
    xtick={0, 1, 2, 3, 4, 5, 6},
    ytick={0, 1, 2, 3, 4, 5, 6},
    x label style={at={(axis description cs:-0.10,-0.1)},anchor=north},
    y label style={at={(axis description cs:-0.25,0.5)},anchor=north},
]

\nextgroupplot[
    hdist style,
    ylabel={$\text{Dist}_\mathcal{H}(\text{LCA}, y)$ : underpred. dist.},
    title=OOD test data,
]\addplot [
    matrix plot*,
    mesh/cols=7,
    point meta=explicit,  
] table [meta=value] {data/dists-inat-ood.dat};

\nextgroupplot[
    hdist style,
    xlabel={$\text{Dist}_\mathcal{H}(\text{LCA}, f(x))$ : overpred. dist.},
    empty colorbar,
    title=ID test data,
] \addplot [
    matrix plot*,
    mesh/cols=7,
    point meta=explicit,   
] table [meta=value] {data/dists-inat-val.dat};
    
\end{groupplot}

\end{tikzpicture}
    \vspace{-0.1cm}
    \caption{Hierarchical distances: iNaturalist19.}
    \label{fig:distances-inat}
\end{figure*}
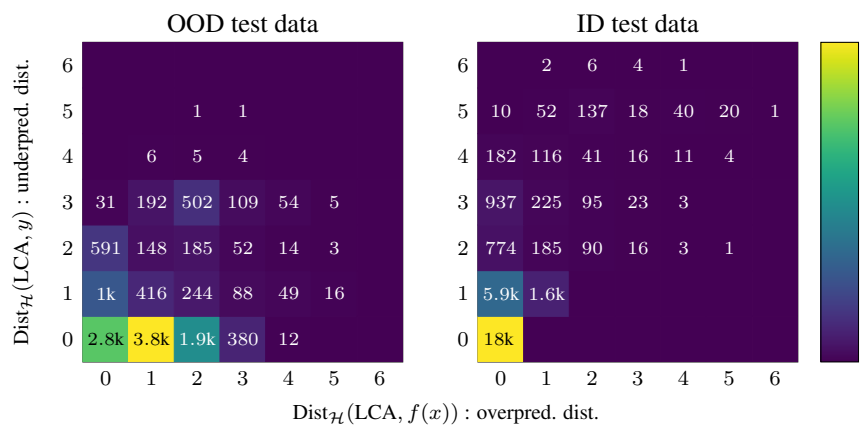
\begin{figure*}[!h]
    \centering
    \begin{tikzpicture}

\begin{groupplot}[
    group style={
        group size=2 by 1,
        vertical sep=1.0cm,
        },
    hdist base,
    width=5 cm,
    height=5 cm,
    xtick={0, 1, 2, 3},
    ytick={0, 1, 2, 3},
    x label style={at={(axis description cs:-0.15,-0.1)},anchor=north},
    y label style={at={(axis description cs:-0.25,0.5)},anchor=north},        
]

\nextgroupplot[
    hdist style,
    ylabel={$\text{Dist}_\mathcal{H}(\text{LCA}, y)$ : underpred. dist.},
    title=OOD test data,
]\addplot [
    matrix plot*,
    mesh/cols=4,
    point meta=explicit,
] table [meta=value] {data/dists-fgvc-ood.dat};

\nextgroupplot[
    hdist style,
    xlabel={$\text{Dist}_\mathcal{H}(\text{LCA}, f(x))$ : overpred. dist.},
    empty colorbar,    
    title=ID test data,
] \addplot [
    matrix plot*,
    mesh/cols=4,
    point meta=explicit,
] table [meta=value] {data/dists-fgvc-val.dat};
    
\end{groupplot}

\end{tikzpicture}
    \vspace{-0.1cm}
    \caption{Hierarchical distances: FGVC-Aircraft.}
    \label{fig:distances-fgvc-aircraft}
\end{figure*}
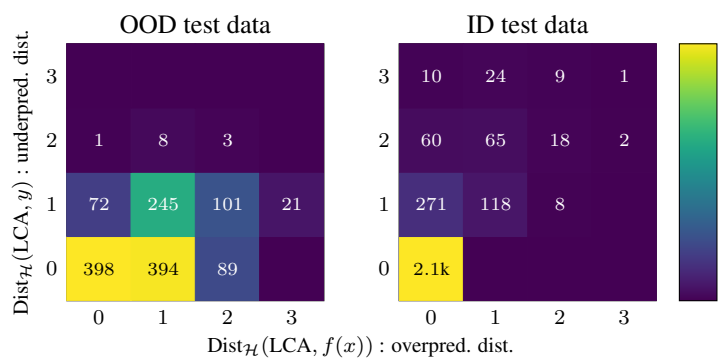
\begin{figure*}[!h]
    \centering
    \begin{tikzpicture}

\begin{groupplot}[
    group style={
        group size=2 by 1,
        horizontal sep=0.7cm,
        },
    hdist base,  
    xtick={0, 1, 2, 3, 4, 5, 6, 7, 8, 9, 10, 11},
    ytick={0, 1, 2, 3, 4, 5, 6, 7, 8, 9, 10, 11},
    x label style={at={(axis description cs:-0.05,-0.1)},anchor=north},
    y label style={at={(axis description cs:-0.15,0.5)},anchor=north},   
]

\nextgroupplot[
    hdist style,
    ylabel={$\text{Dist}_\mathcal{H}(\text{LCA}, y)$ : underpred. dist.},
    title=OOD test data,
]\addplot [
    matrix plot*,
    mesh/cols=12,
    point meta=explicit,       
] table [meta=value] {data/dists-imagenet-ood.dat};

\nextgroupplot[
    hdist style,
    xlabel={$\text{Dist}_\mathcal{H}(\text{LCA}, f(x))$ : overpred. dist.},
    empty colorbar,
    title=ID test data,
] \addplot [
    matrix plot*,
    mesh/cols=12, 
    point meta=explicit,
] table [meta=value] {data/dists-imagenet-val.dat};
    
\end{groupplot}

\end{tikzpicture}
    \vspace{-0.1cm}
    \caption{Hierarchical distances: SimpleHierImageNet.}
    \label{fig:distances-simple-hier-imagenet}
\end{figure*}
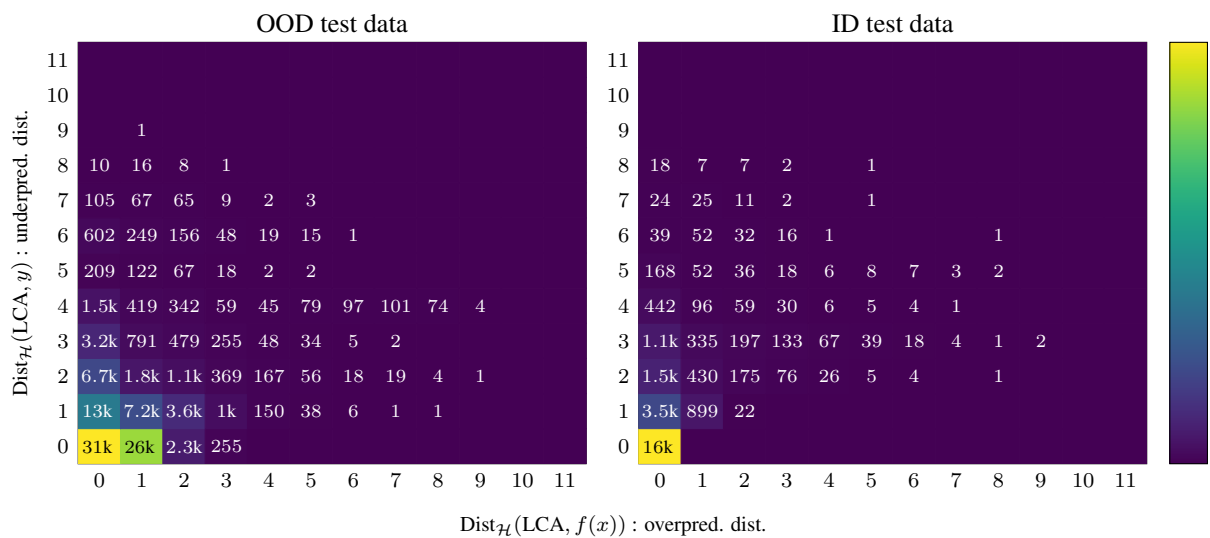

\clearpage

\section{Easy and hard OOD classes}

For a more qualitative evaluation of the performance of ProHOC, we look at which OOD classes get the best and worst performance. Specifically, \cref{tab:easy-vs-hard-ood} shows the top three and bottom three mean hierarchical distances for OOD classes across each test set. We can see relatively large differences between the easy and hard classes for all datasets, with SimpleHierImageNet displaying the largest spread.

\begin{figure*}[!t]
    \centering
    \input{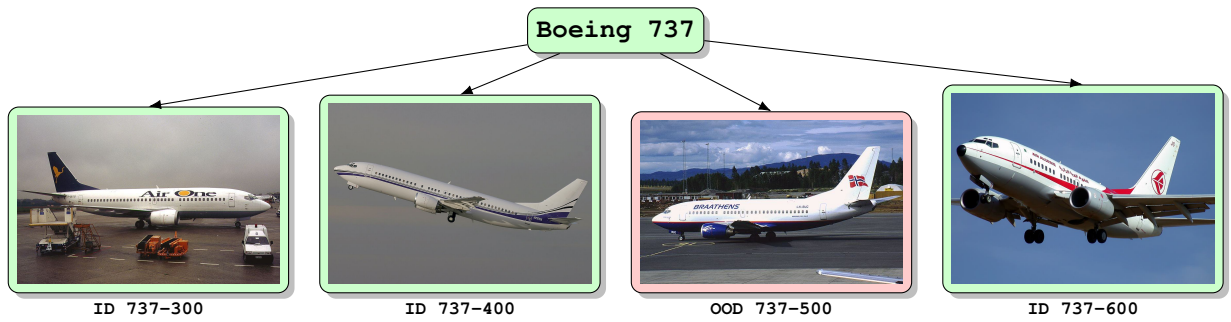} %
    \caption{Easy OOD: FGVC-Aircraft.}
    \label{fig:fgvc-easy}
\end{figure*}

\begin{figure*}[!t]
    \centering
    \input{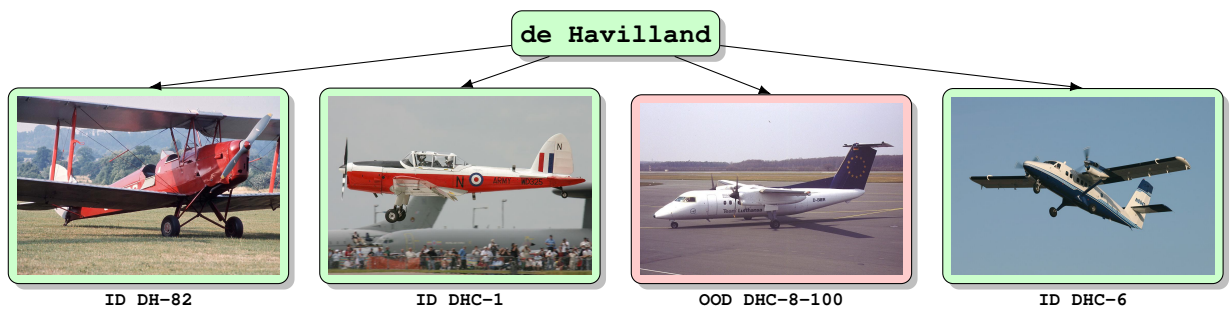} %
    \caption{Hard OOD: FGVC-Aircraft.}
    \label{fig:fgvc-hard}
\end{figure*}

\begin{figure*}[!t]
    \centering
    \input{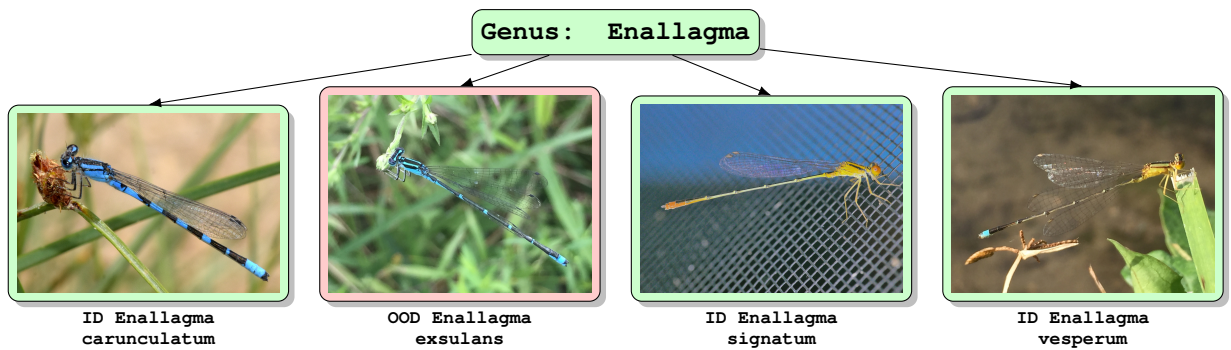} %
    \caption{Easy OOD: iNaturalist19.}
    \label{fig:inat-easy}
\end{figure*}

\begin{figure*}[!t]
    \centering
    \input{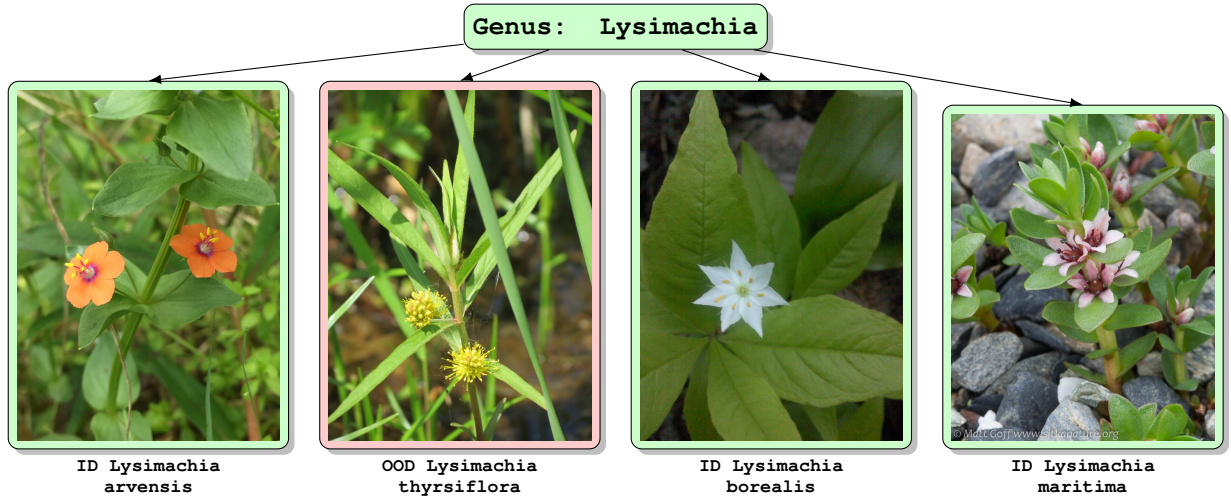} %
    \caption{Hard OOD: iNaturalist19.}
    \label{fig:inat-hard}
\end{figure*}

\begin{figure*}[!t]
    \centering
    \input{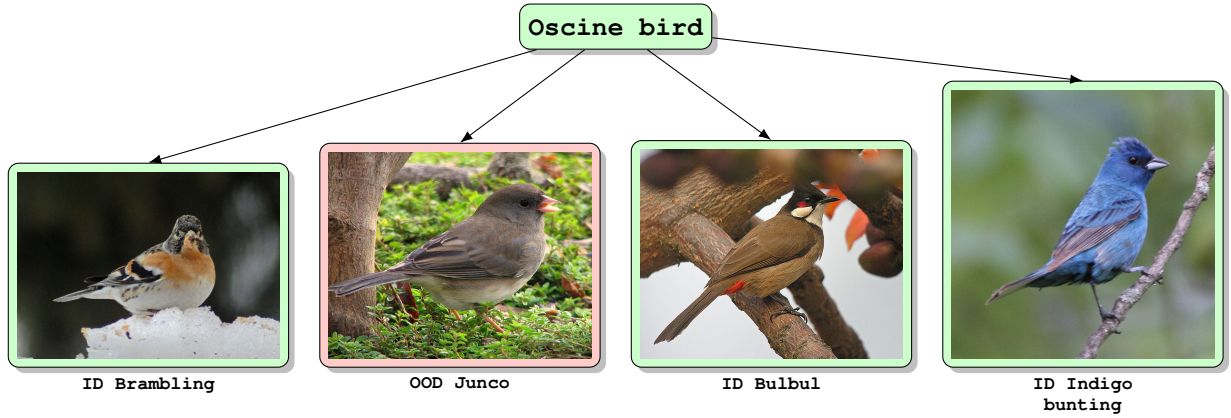} %
    \caption{Easy OOD: SimpleHierImageNet.}
    \label{fig:imagenet-easy}
\end{figure*}

\begin{figure*}[!t]
    \centering
    \input{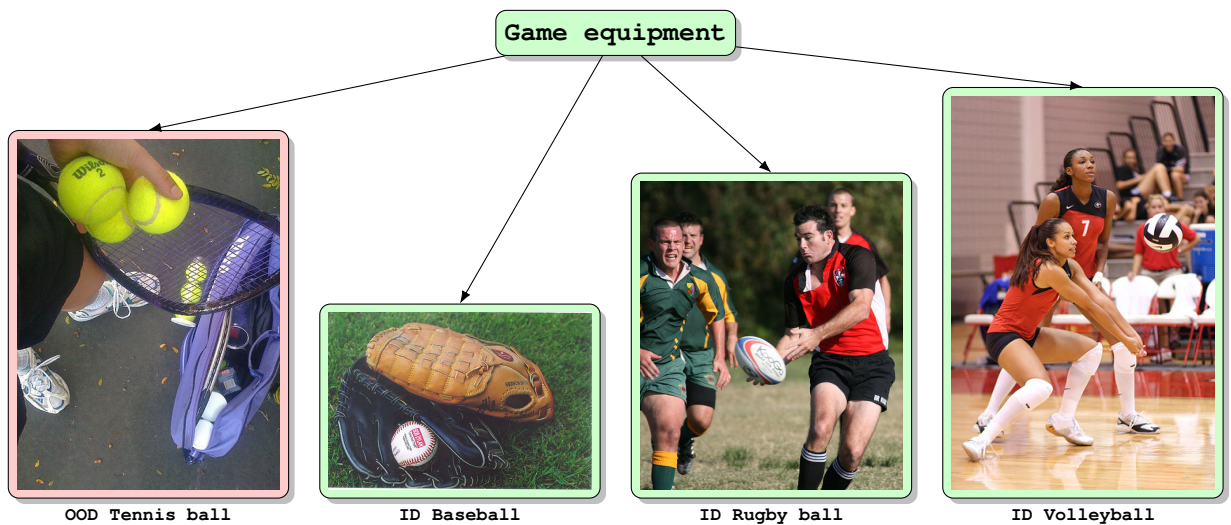} %
    \caption{Hard OOD: SimpleHierImageNet.}
    \label{fig:imagenet-hard}
\end{figure*}

\Cref{fig:fgvc-easy,fig:fgvc-hard,fig:inat-easy,fig:inat-hard,fig:imagenet-easy,fig:imagenet-hard} shows images from the ID and OOD descendants for the top and bottom-performing classes in \cref{tab:easy-vs-hard-ood}. Note that these figures do not display the full hierarchy or all the descendants of the particular nodes. For FGVC-Aircraft, \cref{fig:fgvc-easy} shows that the OOD sample of Boeing 737 closely resembles the ID descendants, making it easy to predict correctly. Conversely, for the hard class shown in \cref{fig:fgvc-hard}, the ID descendants consist of smaller aircraft, whereas the OOD sample is a large passenger plane with few common visual features to the ID descendants, making it challenging to predict accurately. 

For the easy and hard examples of iNaturalist19 shown in \cref{fig:inat-easy,fig:inat-hard} we again see that the ID and OOD descendants in the easy example display strong visual similarities. For the hard example, the flowers differ significantly in color and shape. Additionally, there are many other flower species in the iNaturalist19 dataset, making OOD samples as in \cref{fig:inat-hard} challenging.

SimpleHierImageNet has both the easiest and the hardest classes across all our datasets. The OOD samples for Oscine bird (\cref{fig:imagenet-easy}) get a low mean hierarchical distance of 0.337. We hypothesize that this class is easy because, as in the easy examples above, its descendants share clear visual features, such as body shape, tail, and beak. However, there are also distinct visual features for distinguishing between the descendants, such as colors and patterns, making it easy to identify a sample as part of the group while distinguishing it from the specific ID descendants.

On the opposite end of the spectrum is the Game equipment class (\cref{fig:imagenet-hard}) with a mean hierarchical distance of 4.217. While the model potentially could recognize the round shapes of the balls, the images in these categories tend to be cluttered with various objects and people, making it challenging to identify common features. Additionally, SimpleHierImageNet has, \eg, categories corresponding to clothing that could confuse when there are people in the images.

\begin{table}[!t]
    \centering
    \caption{The top and bottom hierarchical distances per class.}
    \label{tab:easy-vs-hard-ood}    
    \footnotesize
    \begin{tabular}{lr}
 \toprule
 OOD Class & Mean $\text{dist}_\mathcal{H}(f(x), y)$ \\ \midrule
 \multicolumn{2}{c}{\textsc{iNaturalist19}} \\ \midrule
 Genus: Enallagma & 0.43 \\
 Genus: Viola & 0.45 \\ 
 Genus: Aminata & 0.45 \\ \cdashlinelr{1-2}
 Phylum: Angiospermae & 2.17 \\
 Class: Aves & 2.20 \\
 Genus: Lysimachia & 2.66 \\ \midrule
 \multicolumn{2}{c}{\textsc{FGVC-Aircraft}} \\ \midrule
 Family: Boeing 737 & 0.53 \\
 Manufacturer: Douglas Aircraft Company & 0.56 \\
 Family: Airbus A320 & 0.61 \\ \cdashlinelr{1-2}
 Manufacturer: McDonnell Douglas & 1.40 \\
 Manufacturer: Fokker & 1.99 \\
 Manufacturer: de Havilland & 2.02 \\ \midrule
 \multicolumn{2}{c}{\textsc{SimpleHierImageNet}} \\ \midrule
 Oscine bird & 0.34 \\
 Insect & 0.48 \\
 Aquatic bird & 0.51 \\ \cdashlinelr{1-2}
 Cat & 3.23 \\ 
 Kitchen appliance & 3.52 \\
 Game equipment & 4.22 \\ 
 \bottomrule
\end{tabular}
\end{table}

\clearpage
\twocolumn[{
    \begin{center}
        \begin{minipage}{\textwidth}
            \centering
            \captionof{table}{Comparing ProHOC with the ResNet50 backbone and the DINOv2 ViT-L/14 backbone. Results using the ResNet50 backbone are gathered from \cref{tab:main-results}. Excluding the oracle model, the best results are \textbf{boldfaced}.}
            \label{tab:dinov2-results}
            \footnotesize
            \begin{tabular}{l l >{\color{gray}}c >{\color{gray}}c c >{\color{gray}}c >{\color{gray}}c c} %
 \toprule
 & Backbone & $\text{BAcc}_\text{id}$ $\uparrow$ & $\text{BAcc}_\text{ood}$ $\uparrow$ & MixBAcc $\uparrow$ & $\text{BMHD}_\text{id}$ $\downarrow$ & $\text{BMHD}_\text{ood}$ $\downarrow$ & MixBMHD $\downarrow$ \\ 
 \midrule
 & \multicolumn{6}{c}{\textsc{SimpleHierImageNet}} \\ \midrule
 Depth oracle & ResNet50 & 79.7 & 72.5 & 76.1 & 0.82 & 1.05 & 0.93 \\ 
 Depth oracle & DINOv2 ViT & 88.9 & 81.1 & 85.0 & 0.40 & 0.79 & 0.60 \\ \cdashlinelr{1-8}
 ProHOC (CompProb) & ResNet50 & 67.8 & 19.2 & 43.5 & 0.92 & 1.61 & 1.27 \\
 ProHOC (CompProb) & DINOv2 ViT & 85.8 & 18.6 & 52.2 & 0.40 & 1.50 & 0.95 \\
 ProHOC (EntCompProb) & ResNet50 & 62.5 & 30.3 & 46.4 & 0.96 & 1.45 & 1.21 \\
ProHOC (EntCompProb) & DINOv2 ViT & 81.5 & 34.6 & \textbf{58.0} & 0.42 & 1.30 & \textbf{0.86} \\ \midrule
 & \multicolumn{6}{c}{\textsc{iNaturalist19}} \\ \midrule
 Depth oracle & ResNet50 & 72.4 & 75.9 & 74.2 & 0.85 & 0.82 & 0.83 \\
 Depth oracle & DINOv2 ViT & 76.8 & 85.6 & 81.2 & 0.58 & 0.48 & 0.53 \\ \cdashlinelr{1-8}
 ProHOC (CompProb) & ResNet50 & 66.1 & 18.0 & 42.0 & 0.77 & 1.34 & 1.06 \\
 ProHOC (CompProb) & DINOv2 ViT & 72.2 & 23.7 & 47.9 & 0.49 & 1.12 & 0.81 \\
 ProHOC (EntCompProb) & ResNet50 & 57.7 & 35.6 & 46.7 & 0.78 & 1.10 & 0.94 \\
 ProHOC (EntCompProb) & DINOv2 ViT & 60.1 & 49.6 & \textbf{54.9} & 0.54 & 0.82 & \textbf{0.68} \\ \midrule
 & \multicolumn{6}{c}{\textsc{FGVC-Aircraft}} \\ \midrule
 Depth oracle & ResNet50 & 84.7 & 67.6 & 76.1 & 0.49 & 0.67 & 0.58 \\
 Depth oracle & DINOv2 ViT & 85.6 & 61.0 & 73.3 & 0.42 & 0.82 & 0.62 \\ \cdashlinelr{1-8} 
 ProHOC (CompProb) & ResNet50 & 80.1 & 17.1 & 48.6 & 0.41 & 1.25 & 0.83 \\
 ProHOC (CompProb) & DINOv2 ViT & 67.4 & 27.0 & 47.2 & 0.54 & 1.16 & 0.85 \\
 ProHOC (EntCompProb) & ResNet50 & 78.0 & 22.7 & \textbf{50.3} & 0.41 & 1.21 & 0.81 \\
 ProHOC (EntCompProb) & DINOv2 ViT & 55.6 & 44.8 & 50.2 & 0.63 & 0.96 & \textbf{0.80} \\ 
 \bottomrule
 \end{tabular}
            \vspace{1em}
        \end{minipage}
    \end{center}
}]

\section{ProHOC with DINOv2 ViT}

All results in the main paper are obtained from the ResNet50 architecture due to its widespread use in image classification research. ProHOC, however, is architecture-agnostic, requiring only that the architecture produces a probability vector over classes, making it compatible with any SOTA architecture. To demonstrate ProHOC's transferability to other architectures and highlight the performance gains from using a stronger image backbone, we conduct experiments with ProHOC using image features from a frozen DINOv2 ViT-L/14 backbone \cite{oquab2023dinov2}. In this setup, the multi-depth models are replaced with independent MLPs that take DINOv2 features as input. For SimpleHierImageNet and iNaturalist19, we use four-layer MLPs with a hidden dimension of 512 and a batch size of 512. For FGVC-Aircraft, we use single-layer classification heads and a batch size of 128 due to the smaller dataset size. All models are trained for 300 epochs with an initial learning rate of 0.01, decayed to zero at the end of training using a cosine schedule.

The results from training ProHOC with DINOv2 ViT-L/14 are shown in \cref{tab:dinov2-results}. We see big performance improvements compared to the ResNet50 models on SimpleHierImageNet and iNaturalist19, indicating that ProHOC can leverage the capacity of a stronger backbone model. The EntCompProb model again outperforms CompProb with the DINOv2 backbone. On FGVC-Aircraft, the results from ResNet50 and DINOv2 are closer. Interestingly, the ResNet50 oracle model outperforms DINOv2 for OOD classification, suggesting it captures features relevant for OOD predictions that DINOv2 does not. Nevertheless, the overall performance on FGVC-Aircraft remains similar between ResNet50 and DINOv2.

Note that using a pre-trained backbone like DINOv2 for the hierarchical OOD task changes the preliminaries of the problem. Unlike the ResNet50 models, which encounter OOD data only at test time, the DINOv2 backbone has been pre-trained on all our evaluated datasets (including the OOD classes), albeit without labels. This gives DINOv2 an inherent advantage. Therefore, the key takeaway from these results is not a direct comparison between the ResNet50 and ViT architectures, but that ProHOC can benefit from the stronger data representations provided by DINOv2.

\clearpage

\section{ID performance of multi-depth networks}

\Cref{tab:multi-depth-id-performance} shows the ID accuracies of the multi-depth networks used to obtain the results in \cref{tab:main-results}. \Cref{tab:multi-depth-id-performance} also shows the number of nodes assigned to each network. As expected, we see a strong correlation between depth and accuracy. Note that the leaf accuracy for iNaturalist19 differs from the value in \cref{tab:main-results} as \cref{tab:multi-depth-id-performance} shows unbalanced accuracies.

\begin{table}[!h]
    \centering
    \caption{ID accuracies for the multi-depth networks.}
    \label{tab:multi-depth-id-performance}
    \footnotesize
    \begin{tabular}{ccc} 
    \toprule
    Depth $d$ & \# classes at $d$ & Acc \\ \midrule
    \multicolumn{3}{c}{\textsc{iNaturalist19}} \\ \midrule
    1 & 3 & 98.7 \\
    2 & 15 & 97.6 \\
    3 & 58 & 93.1 \\
    4 & 239 & 88.9 \\
    5 & 672 & 78.9 \\    
    6 & 721 & 75.8 \\ \midrule
    \multicolumn{3}{c}{\textsc{FGVC-Aircraft}} \\ \midrule
    1 & 30 & 94.3 \\    
    2 & 63 & 90.3 \\
    3 & 80 & 84.7 \\ \midrule
    \multicolumn{3}{c}{\textsc{SimpleHierImageNet}} \\ \midrule
    1 & 2 & 98.3 \\
    2 & 5 & 97.8 \\
    3 & 43 & 95.9 \\
    4 & 54 & 92.5 \\
    5 & 122 & 88.2 \\
    6 & 240 & 85.9 \\
    7 & 402 & 82.4 \\
    8 & 445 & 80.7 \\
    9 & 471 & 80.2 \\
    10 & 512 & 79.6 \\
    11 & 518 & 79.7 \\
    \bottomrule
\end{tabular}

\end{table}

\section{SimpleHierImageNet}

As discussed in \cref{sec:datasets}, tieredImageNet in its original form is not well-suited for OOD detection in class hierarchies due to several issues. First, it includes sibling classes that do not share common visual features (\eg, \emph{analog clock} and \emph{digital clock}), as well as visually similar classes that are separated by large hierarchical distances (\eg, \emph{laptop computer} and \emph{computer keyboard}). Additionally, it contains many narrow branches, such as parent nodes with only two children (\eg, \emph{duck}), making it difficult to identify common features associated with the parent.

To summarize the desirable characteristics of a hierarchy suited for hierarchical OOD detection, we consider the following criteria:
\begin{itemize}
    \item Siblings should share visual features.
    \item Visually similar classes should be separated by small hierarchical distances.
    \item Internal nodes should have enough children to enable learning of common visual features.
\end{itemize}

With these criteria in mind, we have reorganized parts of the tieredImageNet hierarchy to form SimpleHierImageNet, a hierarchy better suited for hierarchical OOD detection. Specifically, we have pruned internal nodes and moved parts of the hierarchy to satisfy the listed criteria. Additionally, a few classes from tieredImageNet are completely omitted because they lack clear visual connections to other classes in the tree, making them difficult to place within the hierarchy while satisfying our requirements. The omitted classes are
\begin{itemize}
    \item n06359193: website
    \item n03314780: face powder
    \item n04192698: shield
    \item n02840245: binder
    \item n03657121: lens cap
    \item n04423845: thimble
    \item n04507155: umbrella
    \item n03467068: guillotine
    \item n03544143: hourglass
    \item n04355338: sundial.
\end{itemize}

As a result of this curation, we go from 234 internal nodes in the original tieredImageNet to 66 internal nodes in SimpleHierImageNet. The full specification of SimpleHierImageNet is available at \url{https://github.com/walline/prohoc}.

\clearpage

\section{Dataset details}

In \cref{tab:more-dataset-details}, we specify the number of samples in each dataset. The OOD test set for SimpleHierImageNet is large because it is expanded using the OOD classes from the original ImageNet training split. The OOD subsets used in the experiments are listed in \cref{tab:inat-ood,tab:fgvc-aircraft-ood,tab:imagenet-ood} and are also defined at \url{https://github.com/walline/prohoc}. These listed classes represent leaf nodes in the original datasets but subsets of these combine to form OOD data associated with higher levels of the tree.

As a last post-processing step, after defining the ID and OOD subsets, we prune the ID hierarchy by removing nodes with only one child. Specifically, we connect the single child directly to the grandparent and remove the intermediate node. The motivation for this pruning is that we consider it unrealistic for the model to learn the difference between a node and its only child.

\begin{table}[!t]
    \centering
    \caption{The number of samples in the respective datasets.}
    \label{tab:more-dataset-details}
    \footnotesize
    \begin{tabular}{lccc}
    \toprule
     & \makecell{\# ID\\train} & \makecell{\# ID\\ test} & \makecell{\# OOD \\ test} \\ \midrule
    FGVC-Aircraft & 5333 & 2667 & 1332 \\
    SimpleHierImageNet & 665877 & 25900 & 104452 \\
    iNaturalist19 & 156768 & 28078 & 12659 \\
    \bottomrule
\end{tabular}
\end{table}

\newpage

\begin{table}[!t]
    \centering
    \caption{OOD categories for FGVC-Aircraft.}
    \label{tab:fgvc-aircraft-ood}    
    \begin{tabular}{| p{3cm}|} %
    \hline
\begin{lstlisting}[firstline=1, lastline=20]
v-737-500
v-737-700
v-747-400
v-767-200
v-767-300
v-767-400
v-A319
v-A330-200
v-A330-300
v-A340-200
v-A340-300
v-A340-500
v-A340-600
v-Challenger_600
v-DC-6
v-DC-9-30
v-DHC-8-100
v-DHC-8-300
v-E-195
v-Fokker_50
\end{lstlisting} \\
    \hline
    \end{tabular}
\end{table}

\begin{table*}[!h]
    \centering
    \caption{OOD categories for SimpleHierImageNet as WordNet IDs.}
    \label{tab:imagenet-ood}    
    \begin{tabular}{| p{2cm} p{2cm} p{2cm} p{2cm} p{2cm}|} %
    \hline
\begin{lstlisting}[firstline=1, lastline=16]
n01534433
n02088094
n02088238
n02088364
n02088466
n02088632
n02089078
n02089867
n02089973
n02090379
n02090622
n02090721
n02091032
n02091134
n02091244
n02091467
\end{lstlisting} & %
\begin{lstlisting}[firstline=17, lastline=32]
n02091635
n02091831
n02092002
n02092339
n01855672
n02012849
n02093991
n02017213
n02096177
n01688243
n02098105
n01728920
n02099429
n01744401
n02108422
n02106166
\end{lstlisting} & %
\begin{lstlisting}[firstline=33, lastline=48]
n02110185
n02123394
n02397096
n02128925
n02422106
n02481823
n02487347
n01494475
n02643566
n02169497
n02256656
n02279972
n07768694
n03207941
n04542943
n03980874
\end{lstlisting} &
\begin{lstlisting}[firstline=49, lastline=64]
n02883205
n03866082
n02794156
n04548280
n03773504
n09246464
n04515003
n02676566
n07715103
n03394916
n07718472
n03804744
n03642806
n02979186
n04409515
n03179701
\end{lstlisting} &    
\begin{lstlisting}[firstline=65, lastline=80]
n03662601
n03673027
n02814533
n03670208
n03345487
n04560804
n03770679
n04604644
n02793495
n02727426
n03089624
n02825657
n04398044
n04285008
n04370456
n02410509
\end{lstlisting} \\ %
    \hline
    \end{tabular}
\end{table*}

\begin{table*}[!h]
    \centering
    \caption{OOD categories for iNaturalist19 with IDs as specified in iNaturalist19.}
    \label{tab:inat-ood}    
    \begin{tabular}{| p{1.4cm} p{1.4cm} p{1.4cm} p{1.4cm} p{1.4cm} p{1.4cm} p{1.4cm} p{1.4cm} p{1.4cm}|} %
    \hline
\begin{lstlisting}[firstline=1, lastline=34]
nat0996
nat0997
nat0998
nat0999
nat1000
nat1001
nat1002
nat1003
nat1004
nat1005
nat1006
nat1007
nat1008
nat1009
nat0958
nat0963
nat0964
nat0965
nat0966
nat0967
nat0968
nat0969
nat0970
nat0971
nat0972
nat0917
nat0910
nat0668
nat0669
nat0684
nat0688
nat0469
nat0481
nat0486
\end{lstlisting} & %
\begin{lstlisting}[firstline=35, lastline=68]
nat0490
nat0491
nat0492
nat0493
nat0494
nat0495
nat0496
nat0497
nat0498
nat0499
nat0500
nat0501
nat0502
nat0448
nat0454
nat0338
nat0344
nat0792
nat0776
nat0777
nat0778
nat0779
nat0780
nat0781
nat0782
nat0783
nat0784
nat0785
nat0786
nat0787
nat0788
nat0732
nat0762
nat0765
\end{lstlisting} & %
\begin{lstlisting}[firstline=69, lastline=102]
nat0400
nat0881
nat0882
nat0883
nat0884
nat0885
nat0886
nat0887
nat0888
nat0889
nat0890
nat0866
nat0867
nat0836
nat0565
nat0567
nat0621
nat0622
nat0623
nat0628
nat0596
nat0597
nat0598
nat0599
nat0600
nat0601
nat0602
nat0603
nat0604
nat0605
nat0606
nat0607
nat0608
nat0609
\end{lstlisting} &
\begin{lstlisting}[firstline=103, lastline=136]
nat0610
nat0611
nat0612
nat0613
nat0614
nat0615
nat0616
nat0617
nat0618
nat0619
nat0620
nat0583
nat0591
nat0388
nat0363
nat0543
nat0515
nat0644
nat0645
nat0646
nat0647
nat0648
nat0649
nat0650
nat0651
nat0652
nat0653
nat0654
nat0655
nat0803
nat0810
nat0818
nat0830
nat0417
\end{lstlisting} &
\begin{lstlisting}[firstline=137, lastline=170]
nat0431
nat0434
nat0723
nat0891
nat0975
nat0190
nat0166
nat0201
nat0257
nat0258
nat0259
nat0260
nat0261
nat0262
nat0263
nat0264
nat0265
nat0266
nat0212
nat0213
nat0214
nat0215
nat0216
nat0217
nat0218
nat0219
nat0220
nat0221
nat0222
nat0223
nat0235
nat0236
nat0237
nat0238
\end{lstlisting} &
\begin{lstlisting}[firstline=171, lastline=204]
nat0239
nat0240
nat0241
nat0242
nat0243
nat0244
nat0245
nat0246
nat0247
nat0248
nat0249
nat0250
nat0251
nat0252
nat0253
nat0254
nat0255
nat0256
nat0224
nat0225
nat0226
nat0227
nat0228
nat0229
nat0230
nat0231
nat0232
nat0233
nat0234
nat0202
nat0203
nat0204
nat0205
nat0206
\end{lstlisting} &
\begin{lstlisting}[firstline=205, lastline=238]
nat0207
nat0208
nat0209
nat0210
nat0211
nat0318
nat0296
nat0297
nat0298
nat0299
nat0300
nat0301
nat0302
nat0303
nat0304
nat0305
nat0306
nat0282
nat0283
nat0284
nat0285
nat0286
nat0287
nat0288
nat0289
nat0290
nat0291
nat0292
nat0293
nat0294
nat0295
nat0315
nat0012
nat0013
\end{lstlisting} &
\begin{lstlisting}[firstline=239, lastline=272]
nat0014
nat0015
nat0016
nat0017
nat0018
nat0019
nat0020
nat0021
nat0022
nat0150
nat0068
nat0069
nat0070
nat0071
nat0072
nat0073
nat0074
nat0075
nat0076
nat0077
nat0078
nat0079
nat0043
nat0044
nat0045
nat0046
nat0047
nat0048
nat0049
nat0050
nat0051
nat0052
nat0053
nat0054
\end{lstlisting} &
\begin{lstlisting}[firstline=273, lastline=306]
nat0055
nat0056
nat0057
nat0058
nat0059
nat0060
nat0061
nat0062
nat0063
nat0064
nat0065
nat0066
nat0067
nat0032
nat0038
nat0000
nat0004
\end{lstlisting} \\ %
    \hline
    \end{tabular}

\end{table*}

\end{document}